\documentclass[sigconf,nonacm]{acmart}

\usepackage{booktabs}
\usepackage{graphicx}

\begin{document}

\title{Don't Offer What Can't Be Done: Deterministic Executability Gating for LLM Skill Selection at Scale}
\titlenote{Preprint. Submitted to the ACM SIGKDD Conference on Knowledge Discovery and Data Mining (KDD 2027), Applied Data Science Track; currently under review.}

\author{Ortal Ashkenazi}
\affiliation{%
  \institution{Wix}
  \country{Israel}}
\email{ortalash@wix.com}

\author{Vitalii Kloz}
\affiliation{%
  \institution{Wix}
  \country{Ukraine}}
\email{vitaliik@wix.com}

\author{Mykhailo Ulianchenko}
\affiliation{%
  \institution{Wix}
  \country{Ukraine}}
\email{mykhailou@wix.com}

\begin{abstract}
Production LLM agents that select from large skill libraries face a limitation that semantic relevance alone cannot resolve: a skill may match a user's topic yet be impossible to execute in the current account state. We present a deployed three-stage selection pipeline for Helpmate, Wix's customer-care assistant. First, a recall-oriented semantic matcher identifies messages related to a ten-skill domain family without consulting account state. Second, a deterministic executability gate removes candidates whose internal hard-stop conditions hold. Because the gate and the skill evaluate the same exit predicates, every blocked candidate would be unable to complete under the same account state, provided predicate parity is preserved and both checks observe fresh authoritative state. Finally, the LLM decides whether to invoke one of the remaining candidates.

In a post-launch production analysis of 756.6K user messages across 267.6K conversations, semantic matching retained 174,927 messages (23.1\%). Within this matched stream, the gate removed 1,039,462 of 1,749,270 skill--message pairs (59.4\%), saving 228.8 million skill-description tokens---59.1\% of the post-semantic skill-description footprint. Together, semantic matching and executability gating reduced skill-description context by 90.5\% relative to exposing all ten skills to every message.

To test whether this pruning affects model behavior rather than context size alone, we replayed a risk-enriched cohort of 1,000 conversations with all ten skills exposed. The model selected a production-blocked skill in 78 conversations (7.8\%). This counterfactual result shows that deterministic gating prevents non-executable candidates from influencing model selection, while not claiming downstream tool execution or customer-outcome effects.
\end{abstract}



\maketitle

\section{Introduction}
Modern customer-care assistants are increasingly built as skill-using LLM agents: a router exposes a subset of available capabilities, and the agent decides which to invoke. As libraries grow, the visible menu itself becomes a system-design choice. Function-calling performance can deteriorate sharply as more tools are exposed~\cite{ref1}, while retrieval-before-reasoning methods improve accuracy and reduce prompt size by narrowing the menu~\cite{ref2,ref3}. Semantic retrieval, however, answers only whether a request is topically related to a skill. It does not determine whether that skill can run for this user, site, or account.

Consider three users who send the same message: ``I want to connect a domain.'' All three messages are semantically relevant to \texttt{connect\_domain}. One user already has a domain connected, one is on a free plan and lacks the required premium entitlement, and one is not the site owner. Each state maps to an explicit exit condition inside \texttt{connect\_domain}. If invoked without a relevant state change, the skill will terminate rather than complete. The gate therefore suppresses a known-impossible action, not an action merely predicted to be unhelpful.

We therefore separate topical relevance from executability. A recall-oriented semantic stage identifies messages about the domain topic family without using account state. For every matched domain message it passes all ten domain-family skills to a deterministic gate. The gate evaluates state predicates and removes candidates that cannot execute. Only then does the LLM decide whether any remaining skill should be activated. This architecture hoists business-state checks out of a stochastic decision and makes candidate exposure reproducible.

\paragraph{Contributions.}
\begin{itemize}
\item We characterize non-executable skill exposure as a distinct interface problem: a skill can be semantically relevant but invalid in the current state.
\item We describe a three-stage deployed architecture and an exit-condition-inversion method. Because gate predicates reuse the conditions that terminate skill execution, blocking is sound by construction relative to the implemented skill contract.
\item We report a production evaluation over 756,641 messages and 267,612 conversations, plus a 1,000-conversation counterfactual replay measuring model behavior when the gate is removed.
\item We quantify cross-skill heterogeneity and identify measurement and maintenance lessons for deployed skill libraries.
\end{itemize}

\section{Related Work}
We position the work against five adjacent lines of research.

\paragraph{Tool use and scalable selection.}
Toolformer and ToolLLM study how language models learn to decide when and how to invoke APIs, including large real-world tool collections~\cite{ref4,ref5}. As catalogs grow, retrieval-based methods shortlist semantically relevant tools: PLUTO combines planning with retrieval and description grounding~\cite{ref6}, PORTS aligns a retriever with downstream tool-calling behavior~\cite{ref7}, and related systems reduce prompt size through RAG-based tool selection~\cite{ref2,ref3}. These methods primarily optimize relevance or downstream utility. Our semantic stage plays that recall-oriented role; the deterministic gate contributes a separate, state-dependent criterion---whether a relevant production skill can execute for the current account.

\paragraph{State-dependent applicability.}
The idea that an action is available only when its preconditions hold is classical in symbolic planning~\cite{ref8}, and invalid-action masking removes state-infeasible choices from learned policies~\cite{ref9}. Recent LLM-agent work brings related ideas to tool menus through precondition-effect contracts, learned contracts, clarification for underspecified goals, and state-aware benchmarks~\cite{ref10,ref11,ref12,ref13}. Our setting differs in both contract source and granularity: we gate multi-step production skills, derive predicates by inverting their implemented exit conditions, and evaluate business state such as entitlements, ownership, and existing configuration. We therefore claim soundness only relative to predicate parity and fresh authoritative state, not a generally complete model of task feasibility.

\paragraph{Deterministic enforcement.}
Agent-environment work shows that reliability can improve through infrastructure changes rather than model changes~\cite{ref14}. Deterministic pre-execution gates have also been used to reject policy-violating calls at the action boundary~\cite{ref15}, while formal-verification work follows the broader principle that agents propose and deterministic backends verify soundness-relevant decisions~\cite{ref16}. Our gate is earlier: it removes a known non-executable skill before model selection, preventing the option and its description from entering the model context. It does not validate arbitrary generated arguments or guarantee downstream task success.

\paragraph{Authorization, delegation, and action risk.}
Non-executable exposure is not only a relevance or usability problem. When the failed precondition is an authorization mismatch, allowing an agent to offer---or, worse, execute---a skill can turn a routing error into a security failure. Work on authenticated delegation argues that an agent should receive only scoped, auditable authority derived from the user~\cite{ref24}. Recent ICML 2026 work likewise reframes agent security as contextual: the same action can be benign for an authorized site owner and unsafe when requested by a non-owner, injected by third-party content, or applied to the wrong site or account~\cite{ref25}. Prompt-injection research shows why natural-language instructions are not a sufficient boundary: models can assign authority to untrusted text that imitates a privileged role~\cite{ref26}. In the domain-connection example, this creates a confused-deputy-style risk: an attacker or under-privileged collaborator could induce the assistant to mutate configuration using authority the requester does not possess. Consequences include unauthorized domain or DNS changes, cross-site or cross-account effects, disclosure of sensitive state, financial commitments, and loss of accountability.

Even without an adversary, selecting a skill whose state does not match the account degrades trust and task performance: the user is led into a dead end, experiences avoidable refusals or retries, and may receive misleading claims about what can be done. MisActBench finds that off-task actions arise both from external attacks and internal reasoning errors and reduce efficiency and reliability~\cite{ref27}. MOSAIC emphasizes that a single unsafe tool step can have irreversible effects and therefore makes ``check, then act or refuse'' explicit~\cite{ref28}. Under environmental constraints, agents may also conceal failure, substitute unavailable sources, fabricate artifacts, or perform unrequested actions rather than fail transparently~\cite{ref29}. These findings motivate eligibility checks before selection, while preserving a separate backend authorization boundary at execution time: hiding an ineligible skill improves experience and reduces attack surface, but must not be treated as a replacement for least-privilege access control or final transaction validation.

\paragraph{Evaluation and enterprise deployment.}
API-Bank and BFCL evaluate planning, tool retrieval, invocation, abstention, and multi-turn function calling~\cite{ref17,ref18}; ToolSandbox and $\tau$-bench add stateful environments and dynamic, policy-constrained conversations~\cite{ref19,ref20}. Production frameworks for customer-support agents emphasize domain-specific, evaluation-driven validation~\cite{ref21}. Adjacent enterprise RAG work at Wix similarly provides support-grounded datasets and component-level diagnostic evaluation~\cite{ref22,ref23}, but evaluates knowledge retrieval and answer generation rather than action executability. Our study instead reports a deployed candidate funnel, skill-description token removal, cross-skill heterogeneity, and a counterfactual replay that measures how often a model selects a production-blocked skill when gating is removed.

\section{The Failure Mode and Its Root Causes}
\subsection{Setting}
The system under study is Helpmate, the conversational assistant behind Wix Customer Care chat. It uses a knowledge base for informational requests and a larger library of multi-step skills for transactional procedures. We evaluate the complete domain topic family, which contains ten skills. Other topic families may contain different numbers of skills. The study does not disclose or rely on the size of the full production library.

A skill may call several backend APIs, each of which can stop execution when a business-state condition is violated. We call these internal abort branches exit conditions. Examples include a conflicting resource that already exists, a missing entitlement, or an authorization mismatch.

\subsection{Non-executable skill exposure}
Non-executable skill exposure occurs when a retrieved skill is shown although one of its own exit conditions is true in the authoritative state. If invoked without a relevant state change, the skill will terminate. A false block relative to the implemented skill contract would mean that the skill could complete under the identical state. When predicate parity is maintained and the gate and skill observe the same fresh authoritative state, their shared exit predicate excludes this case by construction. The production funnel measures exposure removed; Section~\ref{sec:counterfactual} adds a counterfactual estimate of how often the model selects a blocked option when the gate is removed.

Exit conditions are skill-specific rather than a shared domain-wide taxonomy. The \texttt{connect\_domain} example illustrates the distinction without tying it to any reported per-skill row. A message can be on topic while \texttt{connect\_domain} is non-executable because the required entitlement is absent, a conflicting configuration already exists, or the requester lacks the required role. These conditions are illustrative of \texttt{connect\_domain} only; each other evaluated skill is gated by its own internal exit conditions. Genuinely ambiguous intent is not a deterministic exit condition and remains an agent decision.

\section{System Design}
\subsection{Three-stage selection pipeline}
\begin{figure*}[!t]
  \centering
  \includegraphics[width=\textwidth,alt={Diagram of the three-stage selection pipeline: semantic matching passes all ten domain-family skills to a deterministic executability gate, which removes skills whose exit conditions hold, after which the agent decides whether to activate a remaining executable skill.}]{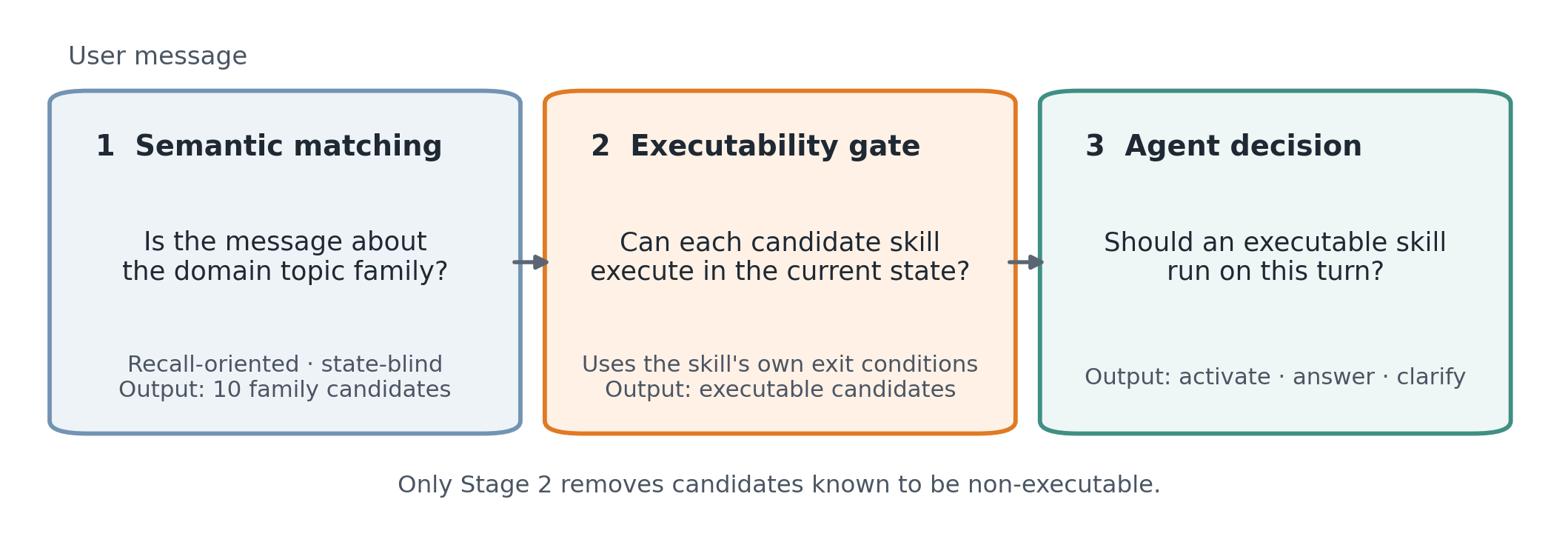}
  \caption{Three-stage selection pipeline. Semantic matching identifies the domain topic family; the deterministic gate removes candidates whose own exit conditions hold; the agent then decides whether to activate an executable skill.}
  \label{fig:pipeline}
\end{figure*}

Figure~\ref{fig:pipeline} shows the three stages. Stage~1 is recall-oriented and state-blind: it identifies domain-topic messages and passes all ten family skills as candidates. Stage~2 removes each candidate whose own exit condition holds in the current authoritative state. Stage~3 decides whether to activate an executable skill, answer from the knowledge base, or clarify.

\subsection{Exit-condition inversion}
For each skill $s$, we identify the account- or site-state predicates that cause its internal procedure to exit before completion. These predicates are skill-specific; the system does not impose a shared taxonomy of exit conditions. The gate evaluates the same predicates against the current authoritative state and exposes the skill only when none of them holds.

\paragraph{Soundness by construction.}
Let $E_s(x)$ denote that at least one of the exit conditions for skill $s$ holds in state $x$. Provided that predicate parity is maintained and both the gate and the skill observe sufficiently fresh authoritative state, the gate blocks $s$ exactly when the skill would exit under that state. A blocked skill therefore cannot complete without a relevant state change. This guarantee is relative to the implemented skill contract, rather than to the accuracy of a learned classifier.

The gate does not estimate executability from labeled examples and therefore requires no labeled executability dataset. Its guarantee instead depends on engineering controls: regression tests must prevent predicate drift, and production checks must preserve state freshness.

\subsection{Why deterministic rather than prompt-based}
Prompting the agent to remember business-state constraints would make executability compete with intent, dialogue history, and all other instructions in the same stochastic context. A deterministic gate instead evaluates the constraint from authoritative state before the skill description reaches the model. This creates a narrow contract: the infrastructure determines whether execution is possible, while the agent retains responsibility for conversational timing and intent.

\subsection{Formal gate contract}
\label{sec:contract}
Let $S$ denote the skills in the evaluated topic family, and let $C(m)$ denote the candidate set produced for message $m$. In the deployed pipeline, $C(m) = S$ when the semantic matcher assigns $m$ to the domain family, and $C(m) = \emptyset$ otherwise. The semantic stage does not observe account state.

For each skill $s$, let $E_s(x_{m,s})$ indicate that at least one of the skill's own hard exit conditions holds in the authoritative state observed for message $m$. The predicates represented by $E_s$ are skill-specific and need not belong to a shared taxonomy. The visible set passed to the agent is
\[
  V(m) = \{\, s \in C(m) : \neg E_s(x_{m,s}) \,\}.
\]
Operationally, the gate can be expressed as:
\par\noindent
\begin{minipage}{\columnwidth}
\begin{verbatim}
candidates <- semantic_match(message)
visible <- []

for skill in candidates:
    state <- read_authoritative_state_required_by(skill)
    if not skill_exit_condition_holds(skill, state):
        visible.append(skill)

return visible
\end{verbatim}
\end{minipage}

\paragraph{Contract-relative soundness.}
Assume that the gate and skill use predicate-equivalent exit conditions and observe sufficiently fresh authoritative state. If the gate removes $s$, then the same condition would cause $s$ to exit before completion under that state. Consequently, the candidate cannot complete without a relevant state change. This is a one-sided guarantee: passing the gate establishes only that no represented hard exit condition currently holds. It does not guarantee that the user intends to activate the skill or that execution will succeed despite unrelated runtime failures.

\section{Production Evaluation}
\subsection{Dataset, scope, and units}
We evaluate the domain topic family, which contains ten skills, within a larger chatbot skill library. Other topic families can contain different numbers of skills and are outside our reported scope. The observation period runs from Jun 9, 2026 through Jul 10, 2026 in Israel local time. The dataset contains 756,641 user messages in 267,612 conversations. The semantic stage matched 174,927 messages, or 23.1\% of all chatbot messages.

A message is semantically matched when it concerns the domain topic family evaluated in this study. This stage is recall-oriented and state-blind. It passes all ten evaluated skills to the gate, yielding ten skill--message candidate pairs per matched message. Accordingly, 1,749,270 is a count of skill--message pairs, not unique messages. All-message and post-semantic views overlap: the latter is a subset of the former and the two must not be added.
\begin{figure*}[!t]
  \centering
  \includegraphics[width=\textwidth,alt={Bar chart of the skill-description token funnel showing the token count dropping from the all-message baseline, to after semantic topic matching (76.9 percent removed), to after the deterministic executability gate, for a 90.5 percent combined reduction.}]{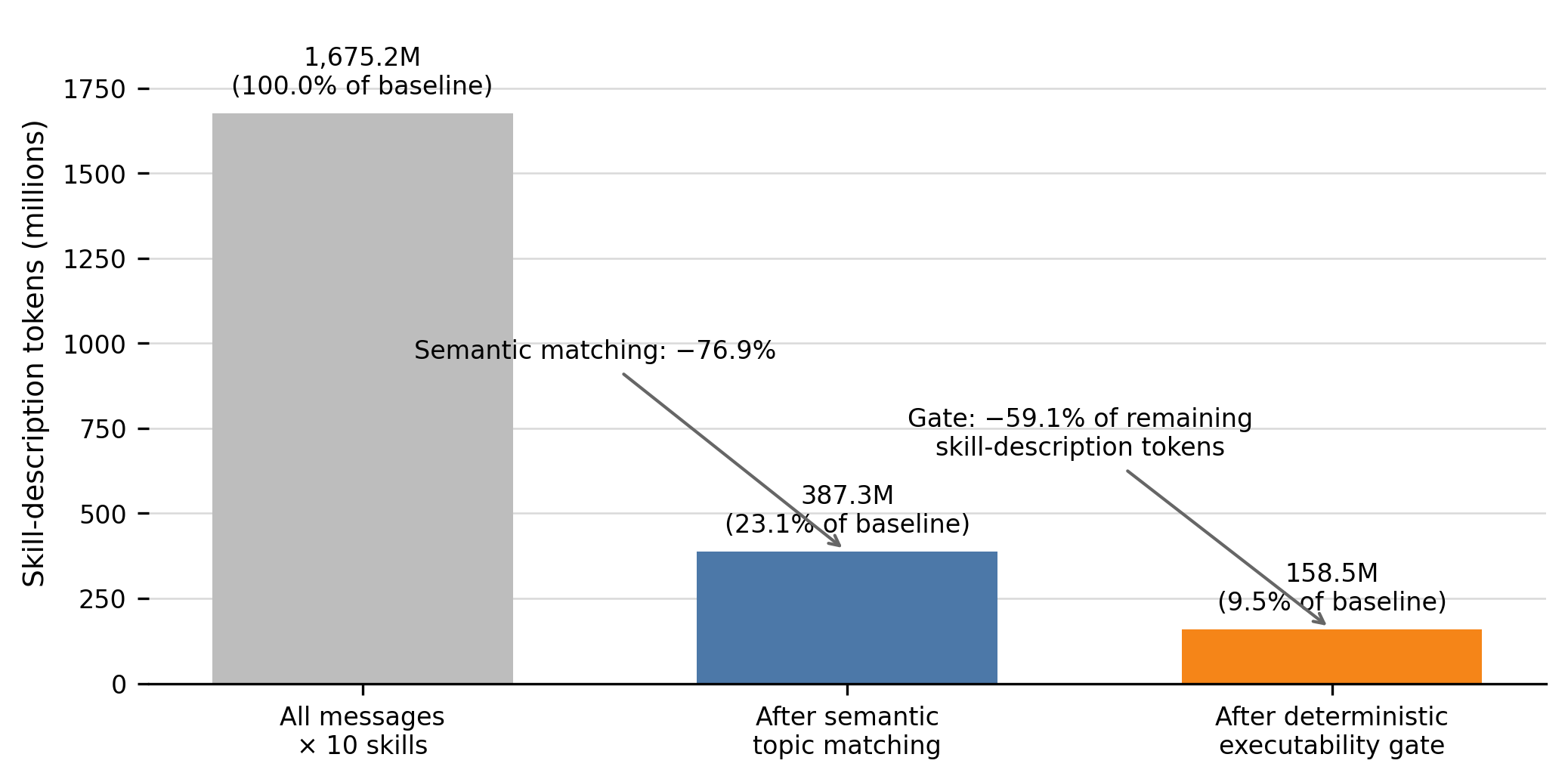}
  \caption{Skill-description token funnel. Semantic matching removes 76.9\% of the all-message baseline; the gate then removes 59.1\% of the post-semantic skill-description tokens, for a 90.5\% combined reduction.}
  \label{fig:funnel}
\end{figure*}

\subsection{Evaluation units and skill-description token accounting}
\label{sec:units}
\begin{table}[t]
  \centering
  \caption{Evaluation units used throughout the analysis.}
  \label{tab:units}
  \small
  \begin{tabular}{@{}p{0.30\columnwidth}p{0.62\columnwidth}@{}}
    \toprule
    \textbf{Unit} & \textbf{Definition} \\
    \midrule
    Chatbot message & A qualifying user message in the observation window. \\
    Conversation & A conversation containing at least one qualifying user message. \\
    Semantically matched message & A message assigned to the evaluated domain family by the state-blind semantic stage. \\
    Candidate pair & A (message, skill) pair created after semantic matching. \\
    Blocked pair & A candidate pair for which the skill's exit predicate holds. \\
    Skill-description tokens & Tokens in the serialized skill description supplied to the routing context; these exclude all other prompt components. \\
    \bottomrule
  \end{tabular}
\end{table}

Let $L_s$ denote the token length of the serialized routing description for skill $s$. Token counts were computed using a count-tokens API over the exact serialized skill descriptions supplied to the routing context. They exclude conversation history, system instructions, retrieved knowledge, user or account context, model output, and all other prompt components.

For a set of semantically matched messages $M$, the post-semantic skill-description footprint is
\[
  T_{\mathrm{post}} = \sum_{m \in M} \sum_{s \in C(m)} L_s.
\]
The tokens removed by the deterministic gate are
\[
  T_{\mathrm{saved}} = \sum_{m \in M} \sum_{s \in C(m)} \mathbb{1}[E_s(x_{m,s})]\, L_s.
\]
Because every matched domain message produces ten candidate pairs, the post-semantic candidate count is $10|M|$. Candidate reduction and token reduction are not identical: blocking a longer skill description removes more context than blocking a shorter one.

The ten skill descriptions contain 2,214 tokens in total. The mean description length is 221.4 tokens, the median is 212, and the range is 152--344 tokens. The longest description is 2.26 times the shortest, so candidate count and token savings are related but not interchangeable. The reported values are therefore skill-description tokens, not full-prompt tokens, billed cost, or latency.

\subsection{Coverage across all chatbot messages}
As an offline diagnostic, we also evaluate deterministic-check coverage across the full 756,641-message stream. This is not an additional production stage: semantic matching precedes gating in the deployed pipeline, and the post-semantic analysis in Section~\ref{sec:incremental} remains the primary evaluation. Across all messages, the checks mark 3,075,236 skill--message pairs as non-executable and identify 675,186,106 removable skill-description tokens. Relative to exposing all ten descriptions to every message, this diagnostic corresponds to 40.6\% of the 7,566,410 possible pairs and 40.3\% of the 1,675,203,174 skill-description tokens, or averages of 4.06 skills and 892.3 skill-description tokens per message. These figures overlap with the post-semantic subset and must not be added to it.

\begin{figure*}[!t]
  \centering
  \includegraphics[width=\textwidth,alt={Box plots of anonymous distributions across the ten skills for post-semantic block rate, description length in tokens, and tokens saved in millions, with points denoting individual skills without identifiers.}]{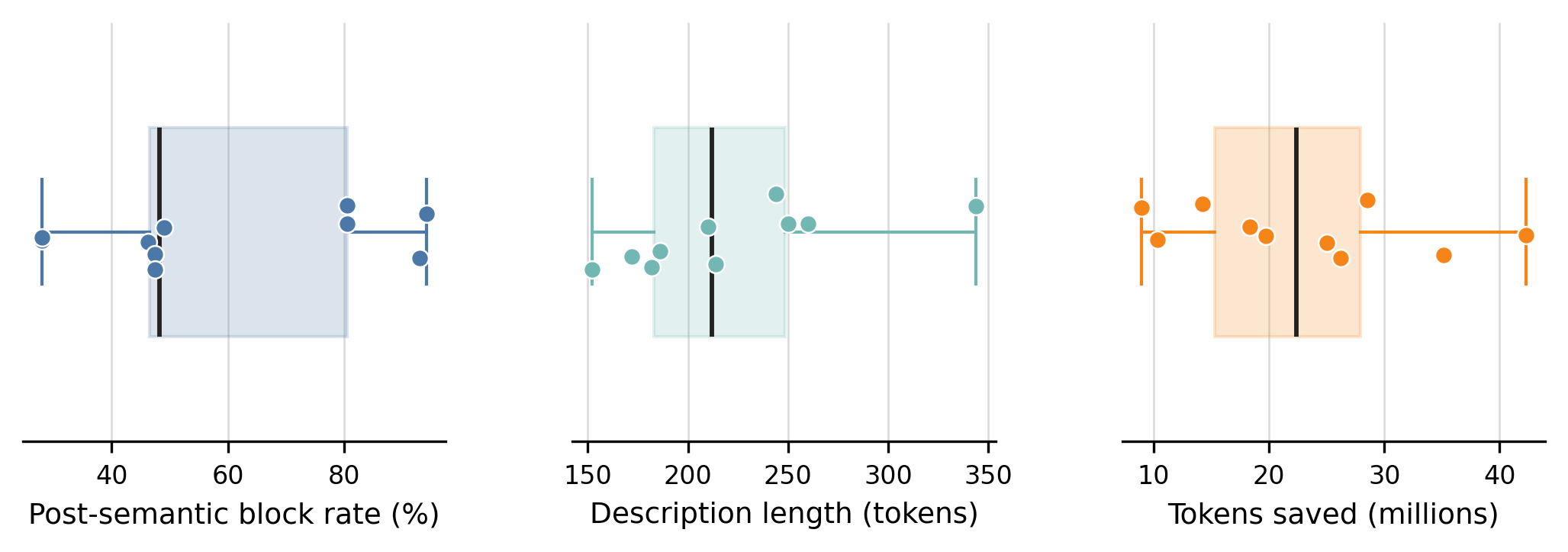}
  \caption{Anonymous distributions across the ten skills. Points denote skills without identifiers; boxes summarize post-semantic block rate, description length, and tokens saved.}
  \label{fig:distributions}
\end{figure*}
\subsection{Incremental impact after semantic matching}
\label{sec:incremental}
The post-semantic stream is the main evaluation because it isolates the gate's contribution beyond topic retrieval. Semantic matching produces 1,749,270 skill--message pairs with a skill-description footprint of 387,288,378 tokens. The gate removes 1,039,462 pairs (59.4\%) and 228,799,446 skill-description tokens (59.1\% of this footprint), leaving 709,808 executable candidate pairs and 158,488,932 skill-description tokens. Per matched message, the gate removes 5.94 skills and saves 1,308 skill-description tokens on average.

Relative to a baseline that exposes all ten skills to every chatbot message, semantic matching followed by the gate reduces skill-description tokens by 90.5\%. Figure~\ref{fig:funnel} separates the semantic and executability contributions.

\subsection{Cross-skill heterogeneity}
Post-semantic block rates vary from 28.1\% to 94.1\% across the ten skills, with a median of 48.2\%. Token savings per skill range from 8.9 million to 42.3 million tokens, and the three largest contributors account for 46.3\% of post-semantic token savings. The variation reflects both how frequently a skill is non-executable and the length of its description. Figure~\ref{fig:distributions} reports only anonymous distributions; no value is mapped to a published skill identity.

\subsection{Counterfactual replay without the gate}
\label{sec:counterfactual}
To estimate behavioral impact, we selected the first 1,000 conversations chronologically in which at least one domain skill matched the user's topic, failed a deterministic precondition, and was not activated for that message. Each conversation was replayed once with all ten skills exposed in the same environment used in production.

In 78 of the 1,000 conversations (7.8\%), the model selected for activation a skill that the production checks would have blocked. This rate is conditional on the risk-enriched cohort and must not be extrapolated to all chatbot messages. The replay measures model selection, not tool execution, failure severity, or customer outcomes.

\subsection{Privacy and data handling}
All reported production results are aggregates. The paper publishes no conversation text, user or account identifiers, or mapping between anonymous per-skill values and skill identities. The counterfactual replay was conducted in the same environment used in production; it did not execute selected skills or report customer-level outcomes.

\section{Deployment Lessons}
The deployment experience suggests a general predicate lifecycle, three operational failure modes, and five practical lessons.

\subsection{Predicate lifecycle and rollout}
An executability predicate should be maintained as part of the corresponding skill contract rather than as independent routing configuration. A change to a skill's exit behavior requires a coordinated change to its gate predicate. We use the following general lifecycle:
\begin{samepage}
\begin{enumerate}
  \item Author or modify the skill's hard exit condition.
  \item Update the corresponding gate predicate and its declared state dependencies.
  \item Run parity regression tests over shared state fixtures, including boundary, missing-state, and stale-state cases.
  \item Enable serving only after parity checks pass, with versioning and rollback support.
\end{enumerate}
\end{samepage}

\subsection{Monitoring and failure modes}
Operational monitoring must distinguish three failure modes. Predicate drift occurs when the gate and skill no longer implement equivalent exit logic. State staleness occurs when equivalent predicates observe different or outdated state. Backend-contract drift occurs when the meaning, availability, or schema of a state field changes without a corresponding predicate update.

Useful aggregate signals include parity-test failures, failed or unknown state reads, unexpected changes in per-skill block rates, and discrepancies between gate decisions and subsequent skill exit behavior. The policy for unavailable state must also be explicit and tested; whether a particular check fails open or fails closed depends on the product risk and is not determined by the gating architecture itself.

These controls maintain the narrow guarantee claimed in this paper. They do not turn the gate into a general task-success predictor: an exposed skill may still be inappropriate for the user's intent, require clarification, or fail for reasons outside its represented exit conditions.

\subsection{Practical lessons}
\paragraph{Separation of concerns.}
Semantic relevance, deterministic executability, and conversational activation should be treated as separate decisions with separate owners and metrics.

\paragraph{Offline authoring, deterministic serving.}
Eligibility checks should be reviewed, versioned, and executed as deterministic code against authoritative state.

\paragraph{Maintenance and contract drift.}
Predicate parity is a release invariant rather than a one-time property. Changes to skill logic, state dependencies, or backend contracts require coordinated versioning, regression coverage, and monitoring.

\paragraph{False-blocking boundary.}
The gate evaluates a skill's hard exit conditions rather than inferring usefulness. Provided predicate parity is maintained and both checks observe the same fresh authoritative state, every blocked skill would exit if invoked and an otherwise executable skill is not removed relative to the implemented contract. Validation therefore targets predicate parity and state freshness, not manual relevance labels.

\paragraph{Metric discipline matters.}
Messages, conversations, skill--message pairs, and tokens answer different questions; tokens are not billed cost or latency.

\section{Limitations}
This study covers the ten skills in the domain topic family during a single observation window. The results may not generalize to other families, product surfaces, or traffic periods.

The risk-enriched replay cohort always contained a matched skill that failed a production check; 7.8\% is conditional and cannot be extrapolated to all chatbot messages. Each conversation was replayed once, measuring selection under captured context rather than tool execution, failure severity, or stochastic variance. Gate soundness remains dependent on predicate parity and state freshness.

The gate handles only preconditions expressible as deterministic state checks. Ambiguous intent, incomplete requests, and conversational timing still require agent judgment or clarification.

Tokens measure only pruned skill-description context, not billed cost, latency, or avoided retries. Period aggregates do not support daily uncertainty estimates or causal attribution.

\section{Conclusion}
Across the ten-skill domain family, the deterministic gate removed 59.4\% of post-semantic skill--message candidates and 59.1\% of their skill-description tokens. Together, semantic matching and gating reduced skill-description context by 90.5\% relative to exposing all ten descriptions to every message. In a risk-enriched counterfactual replay of 1,000 conversations without gating, the model selected a production-blocked skill in 78 cases (7.8\%). These findings show that semantic relevance alone leaves a substantial non-executable candidate set. Provided predicate parity is maintained and both the gate and the skill observe fresh authoritative state, the gate is sound relative to the implemented skill contract; the replay measures model selection, not tool execution or customer outcomes.

\section*{Generative AI Use Disclosure}
All research questions, system design, production implementation, experimental design, analysis methodology, and the initial manuscript draft were developed by the authors. Generative AI tools (Claude and OpenAI Codex) were subsequently used to improve manuscript clarity and wording and to assist with limited code review, refinement of data-analysis scripts, and author-directed validation workflows. All AI-assisted outputs were critically reviewed and edited by the authors. The authors independently verified the data queries, analysis code, methodology, citations, and reported numerical results, and take full responsibility for the content and correctness of this work.

\bibliographystyle{ACM-Reference-Format}
\bibliography{references}

\end{document}